\documentclass[letterpaper, 10 pt, conference]{ieeeconf}  

\IEEEoverridecommandlockouts                              

\usepackage{calrsfs}
\usepackage{graphicx}
\usepackage{gensymb}
\usepackage{booktabs}
\usepackage{multicol}
\usepackage{multirow}
\usepackage{longtable}
\usepackage{graphics} 
\usepackage{epsfig} 
\usepackage{times} 
\usepackage{amsmath} 
\usepackage{amssymb}  
\usepackage{mathtools}
\usepackage{caption}
\usepackage{subcaption}
\usepackage{algorithm} 
\usepackage{algpseudocode} 
\usepackage[colorlinks=true,linkcolor=red]{hyperref}

\DeclareMathAlphabet{\pazocal}{OMS}{zplm}{m}{n}

\title{\LARGE \bf
HiTMap: A Hierarchical Topological Map Representation for Navigation in Unknown Environments
}

\author{Xuecheng Xu, Cheng Wang, Yue Wang and Rong Xiong
\thanks{$^{1}$All authors are with the State Key Laboratory of Industrial Control Technology and Institute of Cyber-Systems and Control, Zhejiang University, Zhejiang, China. Yue Wang is the corresponding author {\tt\small wangyue@iipc.zju.edu.cn}}%
}

\begin{document}

\maketitle
\thispagestyle{empty}
\pagestyle{empty}

\begin{abstract}

The ability to autonomously navigate in unknown environments is important for mobile robots. The map is the core component to achieve this. Most map representations rely on drift-free state estimation and provide a global metric map to navigate. However, in large-scale real-world applications, it's hard to prohibit drifts and compose a globally consistent map quickly. In this paper, a novel representation named, HiTMap, is proposed to enhance the existing map representations. The central idea is to adopt a submap-based hierarchical topology rather than a global metric map so that only a local metric map is maintained for obstacle avoidance which ensures the lightweight of the representation. To guide the robots navigate into unknown spaces, frontiers are detected and attached to the map as an attribute. We also develop a path planning module to evaluate the feasibility and efficiency of our map representation. The system is validated in a simulation environment and a demonstration in the real world is conducted. In addition, the HiTMap is made available open-source. \href{https://github.com/MaverickPeter/HiTMap}{https://github.com/MaverickPeter/HiTMap}
 
\end{abstract}

\section{INTRODUCTION}

Autonomous navigation in unknown environments is a widely studied problem in mobile robots nowadays. It involves researches in online localization, mapping, and planning. In this paper, we focus on the mapping problem aiming to support real-time and large-scale autonomous navigation. In a typical autonomous navigation application, mobile robots need to navigate in partially known environments and incrementally perceive the environment through streaming data provided by onboard sensors. To describe the environment, many map representations are proposed. An intuitive way is to discretize the environment with a set of occupancy grids or voxels \cite{elfes1989using}. Others model the obstacle surfaces with surfels \cite{klingensmith2015chisel, whelan2016elasticfusion, wang2019real} or meshes \cite{piazza2018real, rosinol2020kimera}. All these map representations have their own weakness when it comes to navigation applications. A SLAM system is often used to localize in an unknown environment which exhibit drifts. To correct the accumulated drift error, most map representations need to accommodate corrections and reintegrate all observations into a global map and thus suffering from heavy computation, especially in large-scale environments. The surfaces mapping focuses on merging the texture and cannot be used for navigation directly. 

\begin{figure}[t]
\centering
\includegraphics[scale=0.59]{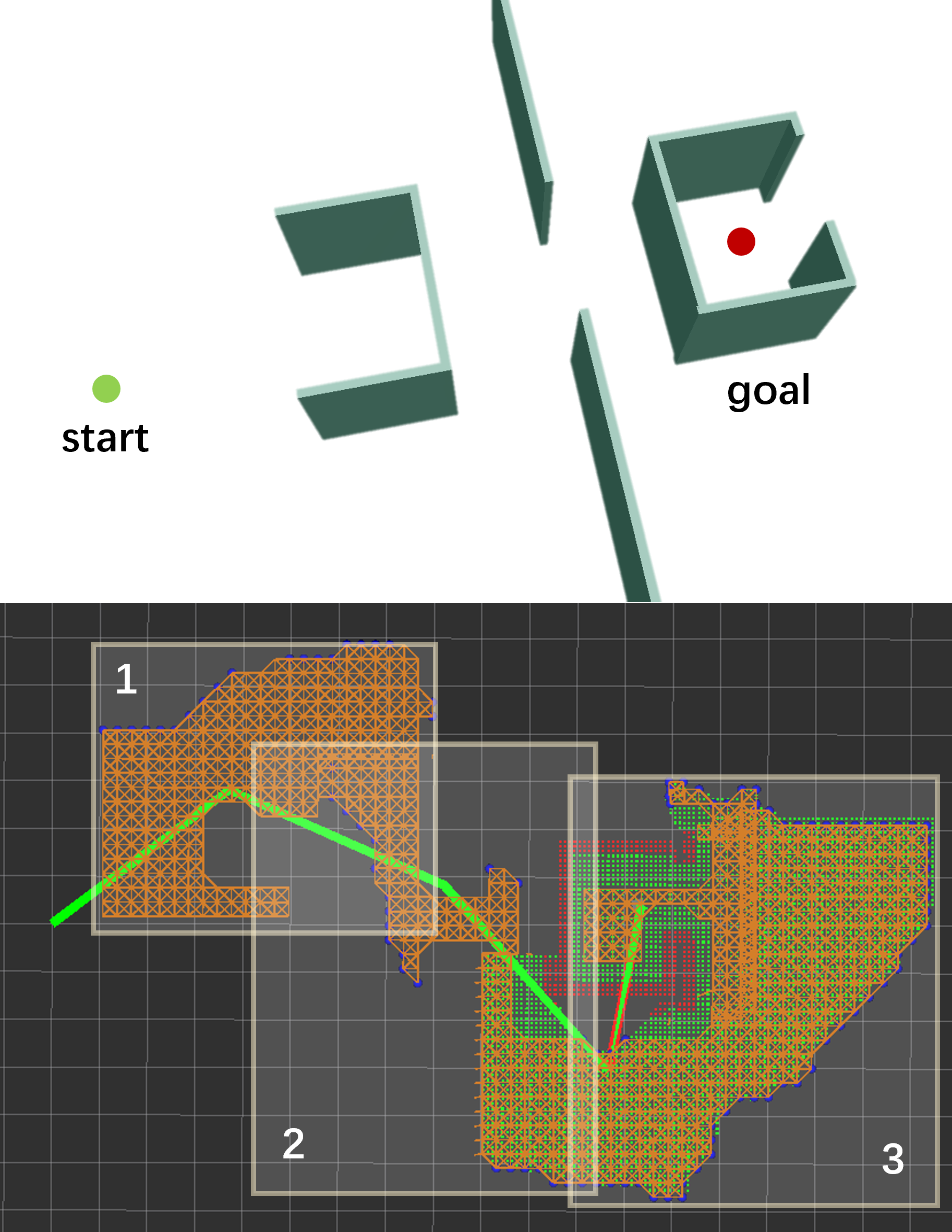}
\caption{We propose HiTMap, a hierarchical topological map representation for navigation in unknown environments. The orange lines are local edges that connect local traversable vertices, the green lines are global edges that connect submaps, blue dots are frontiers. The red points are obstacle points in the local dense map while the green points represent the traversable areas. The transparent area shows the different submaps. In this scenario, HiTMap describes the environment with only one local metric map and neighboring three topological roadmaps.}
\label{fig:Teaser_img}
\vspace{-0.35cm}
\end{figure}

In our preliminary work on occupancy mapping \cite{pan2020gem}, we proposed a real-time globally consistent elevation mapping method called GEM. Attribute to the submap-based 2.5D occupancy representation, GEM achieves better scalability and consistency than  the map representations like OctoMap \cite{hornung2013octomap}. Although GEM is basically satisfied in SLAM systems, there are still impediments to utilize it on large-scale autonomous navigation. The most intractable problem is to achieve global consistency. Although submap-based representation doesn't need to reintegrate all observations, the loop correction and reorganizing a global map are still costly in large-scale environments. Rethinking the purpose of such map representations, the aim is to support exactly optimal global path planning. However, the real-world environment is changing all the time, so optimal global paths are unreliable. With such reflections, we develop an insight that the dense metric information is only effective in the local area.

To address the problems in pure occupancy representations. Recent works utilize rapid-exploring random tree (RRT) \cite{lavalle2001randomized} or its variants to replace the heavy occupancy mapping and improve the efficiency in the exploration task \cite{bircher2016receding, selin2019efficient, schmid2020efficient}. However, the random sampling mechanism within a single iteration may not guarantee the coverage of the free space. Inspired by \cite{xu2021autonomous} to build a full-coverage topological representation, we acquire the insight that full coverage of free space can be achieved by using lightweight sparse roadmap.

Combining the two insights mentioned above, we proposed a hierarchical topological map representation called HiTMap that enables large-scale navigation in unknown environments. Unlike methods using the random tree, we apply a uniform roadmap with the inspiration of Probabilistic RoadMap (PRM) \cite{kavraki1996probabilistic}. The local dense map with traversability information \cite{pan2019gpu} combined with the concept of PRM, HiTMap can achieve comprehensive coverage of traversable areas and generate a reusable sparse topological roadmap. To improve efficiency, an incremental topological mapping mechanism is proposed. Along with the sensor's perception area, a local area consists of several submaps is formed by reusing the stored neighboring roadmap. Also, a coarse topology that is similar to the pose graph in SLAM systems is generated to provide hierarchical large-scale perception. Note that, unlike the pose graph in SLAM systems, a connectivity validation mechanism is proposed to ensure the correctness of the topology relations. As we only store the metric information in vertices of global topology, we just need to add an edge to the existing topology and change the coordinates of corresponding submaps when loop closing is performed.

The contributions of this work are:
\begin{itemize}
   \item A hierarchical topological map representation, HiTMap, is proposed for efficient navigation in large-scale environments with state estimates drift. 
   \item A reusable sparse topological roadmap that guarantees the coverage of free space is proposed. Accordingly, a loop connectivity validation based on this roadmap is implemented for correct topological mapping.
   \item Evaluations on a simulation environment and a demonstration on real-world are conducted. We make our map representation HiTMap available as open source.
\end{itemize}

\section{RELATED WORK}

\subsection{Map Representations}

\subsubsection{Metric Representations}

Metric representations models environments with dense geometrical entities in a specific reference frame. To achieve metric representation, quantization and discretization are necessary. The predominant metric representations are occupancy grid representations in which each grid models an attribute of this area \cite{elfes1989using}. The attribute of a grid can be defined as discrete occupancy conditions or probability or even signed distance to the surface of the closest obstacle \cite{curless1996volumetric, newcombe2011kinectfusion}. In some applications, the polygon is also used to represent known free space \cite{ge2011simultaneous, gonzalez2002navigation}. However, heavily relied on precise location information, metric representations suffer from state estimation drift which is ubiquitous in the real world. Wrong estimations would lead to the inconsistency of the map and further influence the performance of navigation. A solution to the inconsistency is adopting a loop correction mechanism. Submap-based map representations can deal with the drift when there is loop closure happens \cite{pan2020gem} but the price of introducing such a mechanism is high. The correction of submaps rather time-consuming or computational-consuming.

\subsubsection{Topological Representations}

Topological representations models environments with a graph. Places are often regarded as vertices and relations between places are represented by edges. Pure topological maps are scarcely seen in navigation as there is not enough metric information. Typical usage of topological representation is the pose graph in SLAM. Each vertex represents a submap coordinate and edges are simply the distance between vertices. Such representations just provide trajectory information and can be hardly used for autonomous navigation. In recent years, works on exploration tasks often adopt topological map representations as it's suitable for quick planning algorithms like Rapid-exploring Random Tree.

\subsubsection{Hybrid Representations}

Hybrid representations combine advantages of topological and metric representations. Metric maps maintain detailed global location information and have trouble being lightweight. Topological maps, though, are extremely lightweight but lack metric information to navigate. The intuitive combination is to locally build a metric map and navigate in the global topological graph. With this concept, some previous works develop systems for long-term mapping \cite{tang2019topological} and subterranean challenges. However, relying on the place recognition module these methods do not check the connectivity of neighboring submaps, thus lacking traversability information among looped submaps.

\subsection{Navigation in Unknown Environments}

Navigation in unknown environments have mainly two criteria. First, a robot needs to decide where to move next. The second is to quickly perceive the unknown space. There are mainly two categories of planning approaches for such tasks. A prevalent way is frontier-based planning. The frontiers are free areas adjacent to the unknown areas. A common idea is to use frontiers as guidance to navigate robots into unknown areas. However, purely follow the nearest frontier repeatedly may leads to unnecessary back-and-forth movements \cite{cieslewski2017rapid, selin2019efficient}. To address this kind of method, sampling-based methods are proposed. Next-best-view (NBV) \cite{bircher2016receding} planner decides the next view to move according to a utility function. With this concept, graph-based path planning (GBP) \cite{dang2019graph} utilizes Rapidly-exploring Random Graph to choose the next goal with the highest score. Recent work Efficient autonomous exploration planning (AEP) \cite{selin2019efficient} adopts NBVP for local exploration and frontier for global navigation to improve the performance of exploration. The methods mentioned above can achieve competitive performance in simulation and small scale environments, but they failed to meet the large-scale demands. They rely on accurate state estimations and have trouble dealing with drifts.

\section{PROPOSED METHOD}

\begin{figure}[t]
\centering
\includegraphics[scale=0.4]{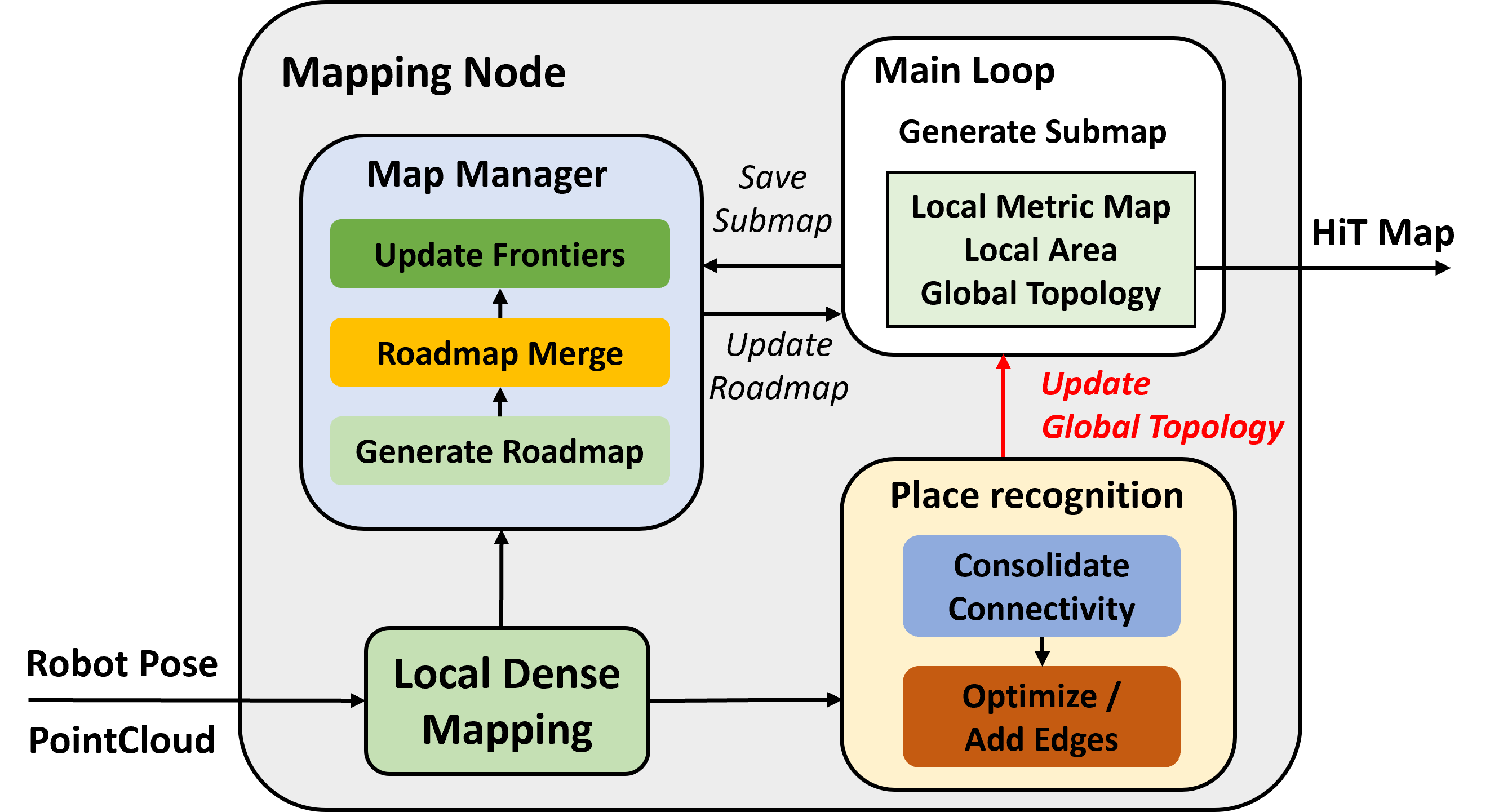}
\caption{HiTMap generation overview. The roadmap is incrementally updated whenever a new observation is received. When there is a loop detected, the previous roadmap can be accordingly updated by simply change the center location of submaps.}
\label{fig:overview}
\vspace{-0.35cm}
\end{figure}

The task of navigation in unknown environments involves works on mapping and planning modules. In this paper, we focus on the map representation that has long been a problem in large-scale real-world navigation applications. Recent map representations used in this task suffer from heavy computation on pose corrections. To address these problems, the main idea of our paper is to propose a map representation that is lightweight and can easily accommodate pose corrections. The HiTMap generation process is shown in Fig. \ref{fig:overview}. We also implement a simple hierarchical planner for our map representation and achieve the navigation in unknown environments.

\subsection{Local Mapping and Planning}

The robot-centric local dense map is important for safe navigation as it provides enough information to avoid the collision. In the local map, we can reasonably assume that state estimation drift is low and we can accumulate sensor data into a dense occupancy map. Here we adopt our preliminary works to simultaneously update the local elevation map and calculate traversability. The main idea is to use height variance and gradient of a grid to calculate its roughness and slope, thus estimate its traversability. Mature local planning algorithms can be easily deployed on this local map.

With the traversability information, we first inflate obstacles to eliminating the influence of robots configuration and further uniformly sample the traversable areas and generate a local topological map. 

The local map maintains the newest environment information thus guarantee safe navigation. Because the local map has fixed bounds and sensors are streaming at a uniform rate, the local mapping process has bounded memory and computation consumption.

\subsection{Local Area}

\begin{figure}[t]
\centering
\includegraphics[scale=0.5]{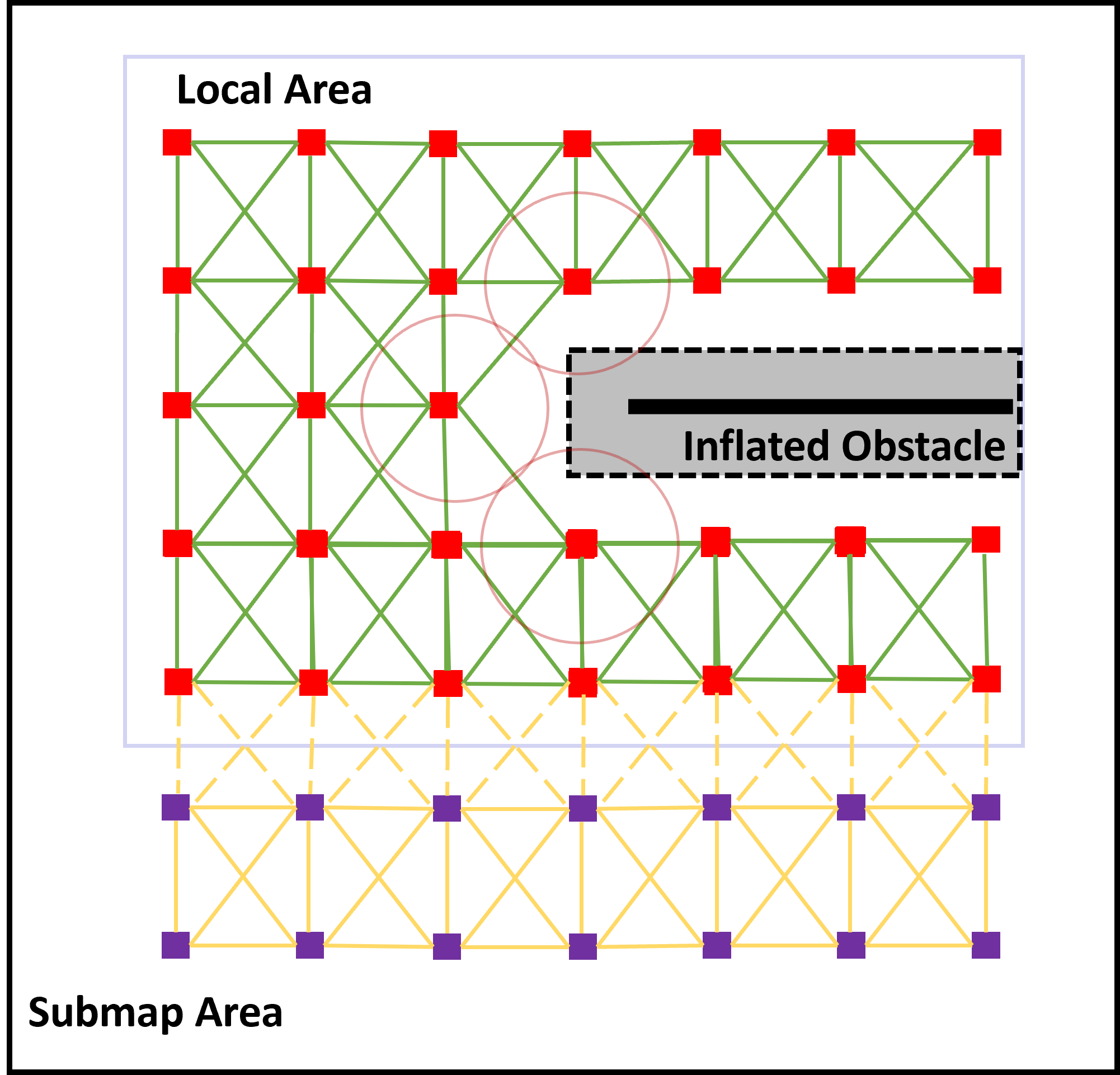}
\caption{The local topological map is incrementally merged into a submap. The purple dots are vertices in a submap that won't update while the red dots are up-to-date vertices based on the local metric map. Green lines are the edges between local vertices. Yellow lines are incremental edges.}
\label{fig:submap}
\vspace{-0.35cm}
\end{figure}

The submap mechanism is often adopted to the map representations that can accommodate pose corrections. Following the step in \cite{pan2020gem} the information that last seen in the local map is merging into the submap. In the topological map, the last seen vertex is incrementally added into the topological submap. The incremental generation of the topological submap can be found in Fig. \ref{fig:submap}.

Based on submaps, we further use local area to provide topology around robots as in \cite{schmid2021unified}. It consists of a few submaps that have connections with the current one. The combination of neighboring submaps extends the robot's vision to the visited places and thus provides an optimal path in the local area. Unlike generating a global metric map as other map representations, local areas only need a bounded lookup and merging consumption.


\subsection{Loop Validation and Correction}

\begin{figure}
\centering
\includegraphics[scale=0.5]{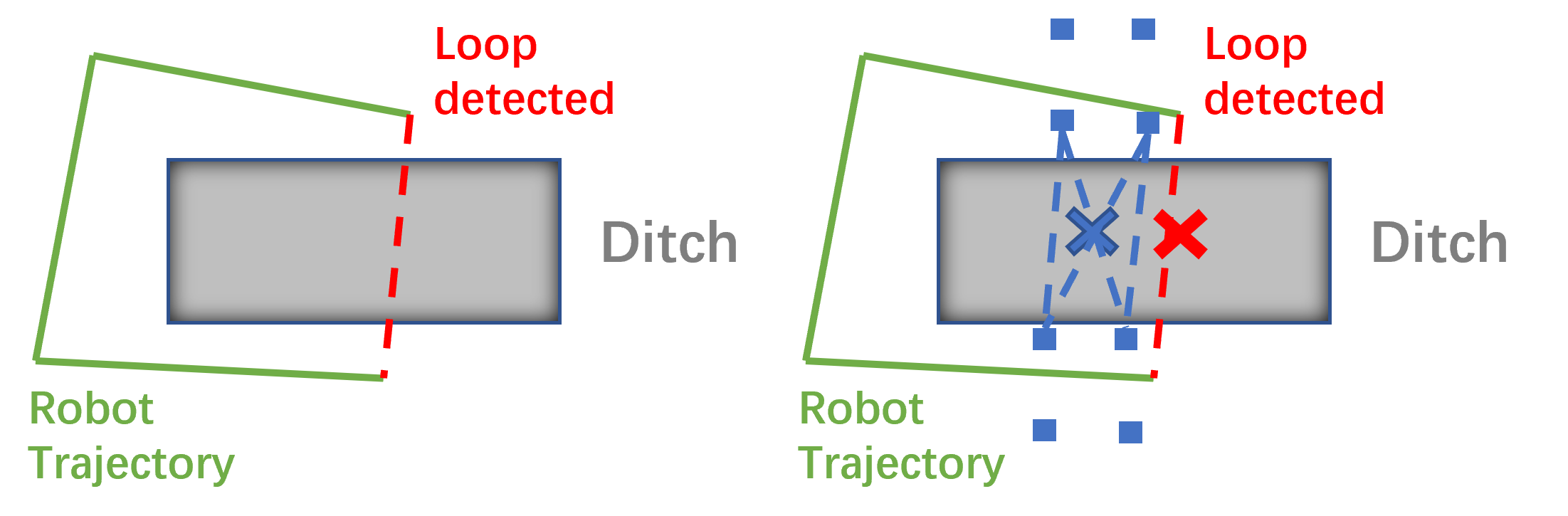}
\caption{A demonstration of the loop validation. The detected loop can not guarantee topological connectivity.}
\label{fig:loopValidate}
\vspace{-0.35cm}
\end{figure}

Place recognition is important in SLAM systems to correct accumulated drifts. It's often implemented by matching two measurements, if matched, two vertices of the pose graph are connected and further optimization process will correct the pose graph. Many approaches also take inspiration from graph SLAM and build a graph with active loop closing \cite{lee2021real}. However, the motion of aerial robots is more flexible than ground robots so that the loop closing result can not equal topological connectivity. An example is shown in Fig. \ref{fig:loopValidate}. To tackle this case, looped submaps should be validated using their topological relations. If the connectivity is checked, two submaps are connected and merged to form the local area.

The pose graph in the SLAM system is corrected after optimization so is our HiTMap. Without the global map, the pose correction of HiTMap is simply changing the coordinate of each submap.

\subsection{Frontiers Update}

Frontiers are boundaries between the known and unknown parts of a map. They can guide the robot to explore the unknown environment and thus achieve navigation in unknown environments. In the HiTMap, frontiers are detected in every submap and saved as an attribute of vertices in the roadmap. As the robot is always in the traversable area, we implement a simple frontier detection module with the idea of Wave Front Detection (WFD) \cite{keidar2014efficient}.

There are many types of frontiers that appeared in different layers in our HiTMap. The first one is the local frontiers detected in the local metric map. These frontiers can be frontiers or vertices connected with the incremental roadmap or even the previous submap. The second type of frontiers is detected in the incremental area of a submap. This kind of frontier is also ambiguous as it may be connected to the previous submap. To tackle this ambiguity, we check the frontiers in the merging step of the local area.

\subsection{Global Planner}

The information provided by the HiTMap contains two parts. The first is the roadmap graph and frontiers of the local area, and the second is the graph and frontiers of the global topology. The first step of global path planning is to add the robot's current position and the goal position into the HiTMap. There are two situations in practical use. The robot is always surrounded by the roadmap, but the goal might be in the known roadmap or not. Accordingly, global path planning can be divided into two modes, the exploration mode, and the backtracing mode. The exploration mode is mainly used to navigate toward the goal that is not in the known roadmap, and the backtracing mode is used to navigate in the established roadmap.

\subsubsection{Backtracing Mode}

If the goal is in the local area, the planning is actually backtracing. The backtracing using roadmap is actually the planning in a graph. With the topological local roadmap, this can be easily achieved by A* algorithm \cite{hart1968formal}.

\subsubsection{Exploration Mode}

If the goal is in the unknown space, the exploration mode is activated.  The first step is to calculate a cost function of each frontier and find the best one as a global waypoint in the topology. The criterion of the global waypoint is formed as Eq. \ref{eq:waypoint}. It has minimal cost value which consists of two parts. One is the distance, the nearest point has the minimum score. The second part is the shift score, it ensures the chosen frontier won't change drastically which often happens in an intersection. And then, simply using the A* algorithm, a global topological path is generated. Since the local area contains a merged roadmap of neighboring submaps, the next global waypoint is actually the nearest connected submap vertex. To navigate towards this waypoint, the nearest frontier is picked. So far, the global path has been converted to series of waypoints and finally connected to the local area. The planning in the local area is degrade into the backtracing problem and can be easily solved by any existing planner.

\begin{equation}
   F_{waypoint} = \arg\min U(F_i)
   \label{eq:waypoint}
\end{equation}
where the $F_i$ indicates to the $ith$ frontier. The function $U$ is defined below.

\begin{equation}
   U(F_i) = W_d * dist(F_i, goal) + W_l * \pazocal{C}(F_i)
   \label{eq:utility}
\end{equation}
where the $W_d$ and $W_l$ are two weight factor. $dist(\dot)$ is the euclidean distance between two points. The function $\pazocal{C}$ is defined below.

\begin{equation}
   \pazocal{C}(F_i) = dist(F_{last} - F_i)
   \label{eq:changeCost}
\end{equation}
where the $F_{last}$ indicates the last chosen position.

\section{IMPLEMENTATION DETAILS}

We use the GEM \cite{pan2020gem} package as the mapping baseline. Based on the original mapping method, we further modified it to provide costmap. The modified version of GEM can be found in this link \footnote{https://github.com/ZJU-Robotics-Lab/GEM}. The pointcloud map generated by GEM is converted into local and global costmaps. The mature DWA planner \cite{fox1997dynamic} and A* global planner \cite{hart1968formal} integrated in ROS is used for planning.

\setlength{\tabcolsep}{1mm}{
\begin{table}[t]
\caption{Simulation (top) and real-world (bottom) experiment configurations.}
\centering
\begin{tabular}{llll}
\hline
local map size & 5m & submap interval & 5m \\
& 15m & & 10m \\
local map resolution & 0.1m & roadmap sample interval & 0.3m\\
& 0.2m & & 0.8m \\
sensor range & 5m & sensor type & depth camera\\
& 15m & & lidar \\
environment scale & 15m x 20m & & \\
& 60m x 70m & & \\

\hline
\end{tabular}
\label{tb:simConfiguration}
\end{table}
}

\begin{figure}
\centering
\includegraphics[scale=0.47]{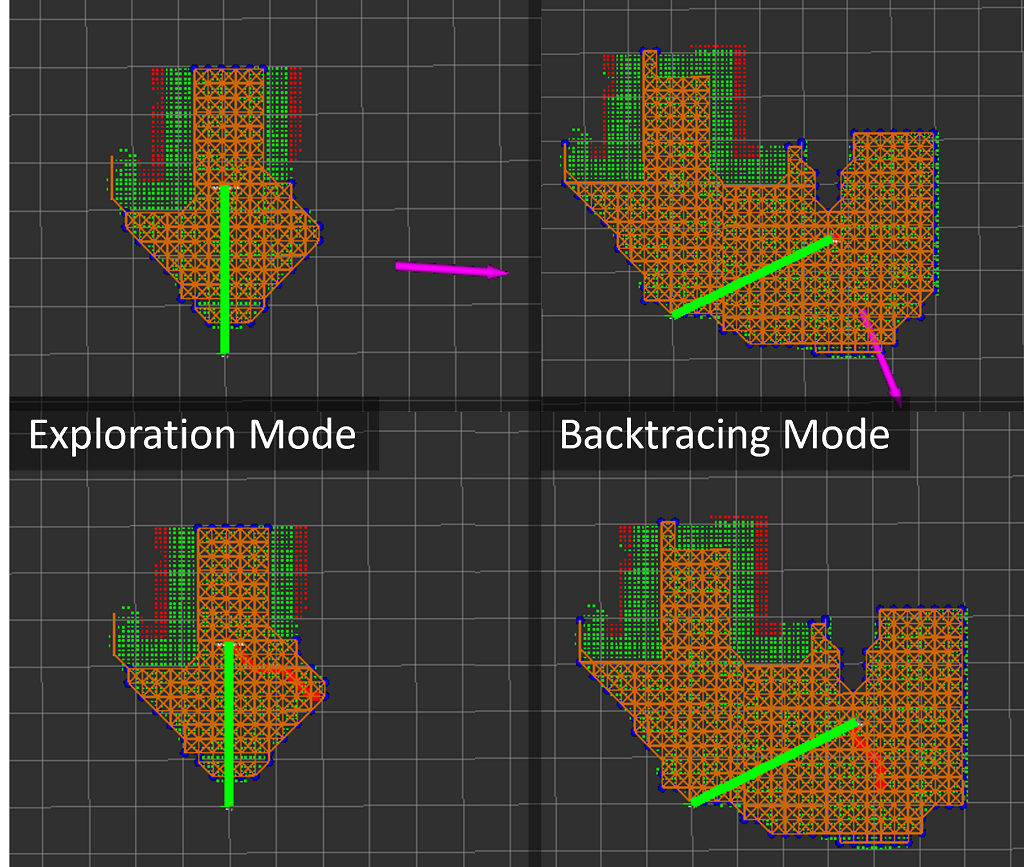}
\caption{A demonstration of two modes in the simulation environment. The exploration mode aims to find the best frontier to the goal and the backtracing mode is to find the best path (red line) in the given roadmap.}
\label{fig:twoMode}
\vspace{-0.35cm}
\end{figure}

\begin{figure}[t]
\centering
\includegraphics[scale=0.47]{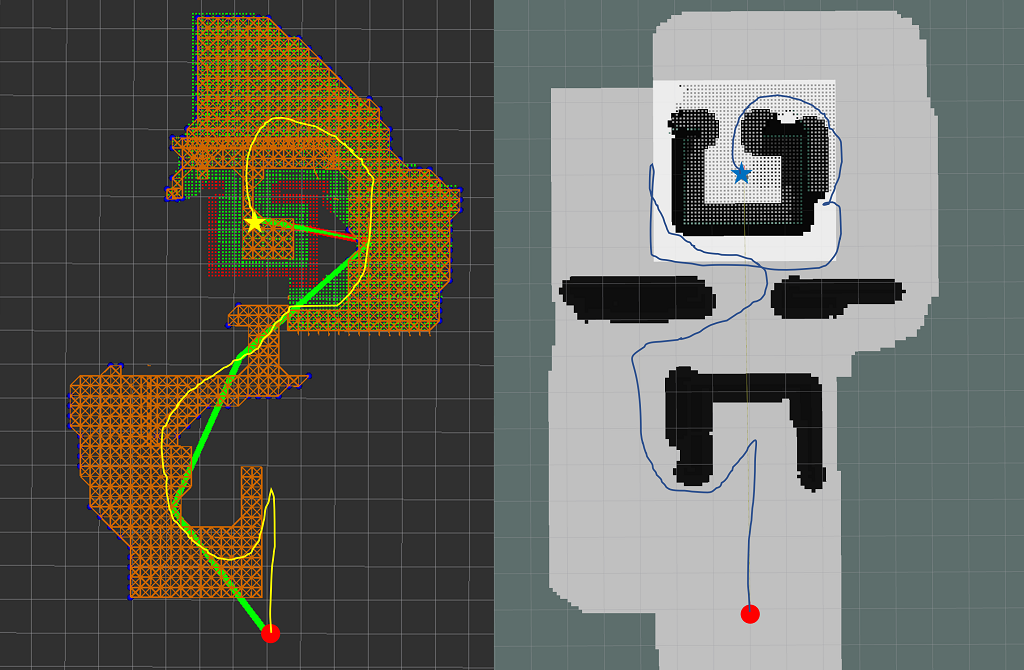}
\caption{Experiment results on the simulation environment. The orange lines on the left are the roadmap connection of the local area. The green line represents the global topology. The red circle is the start point and the yellow star is the goal. On the right, the global costmap is shown.}
\label{fig:simulation}
\vspace{-0.35cm}
\end{figure}

\begin{figure}[t]
\centering
\includegraphics[scale=0.42]{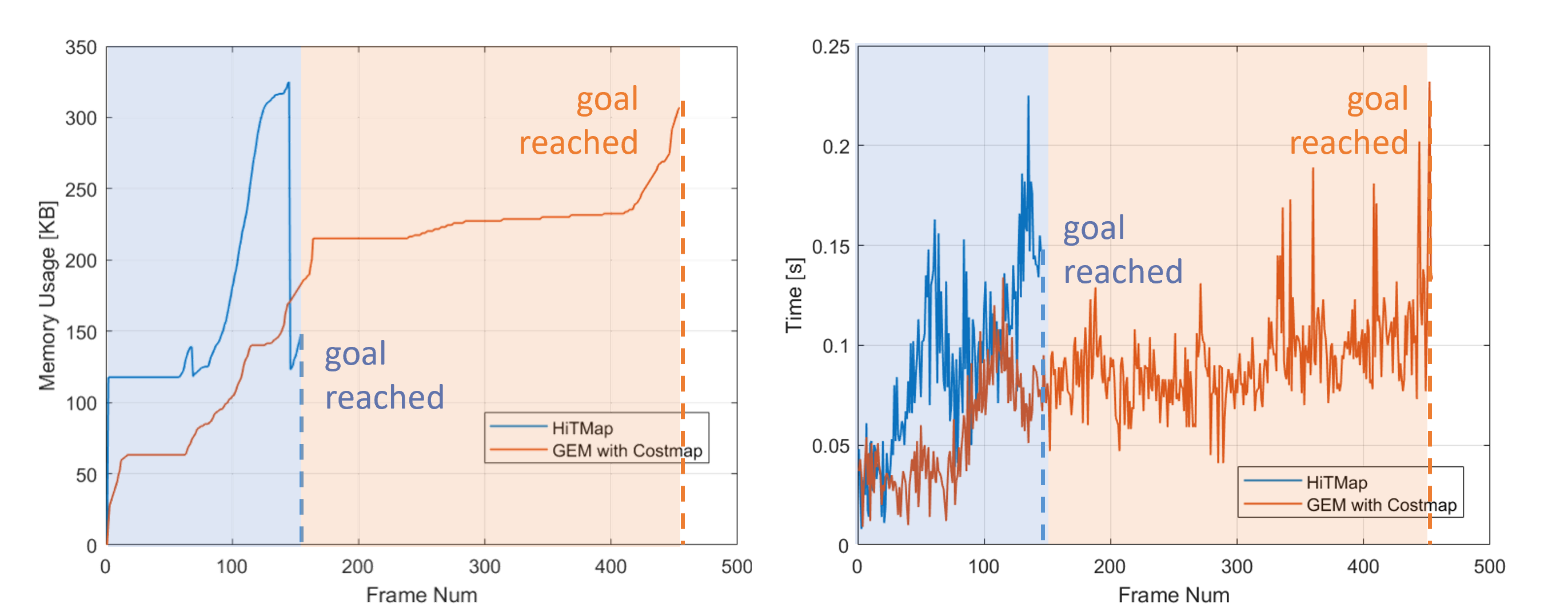}
\caption{The memory usage and time consumption trend in the simulation experiment.}
\label{fig:simResult}
\vspace{-0.35cm}
\end{figure}

\section{EXPERIMENTS}

The experiments are all conducted with ROS running on a personal computer with AMD Ryzen 3700X and NVIDIA RTX 2060 Super. All experiments are to evaluate the feasibility of the HiTMap for navigation in unknown environments and its efficiency. We choose the most recent and competitive metric map representation, GEM \cite{pan2020gem}, as our benchmark. The configuration of experiments are shown in TAB.\ref{tb:simConfiguration}. The distance-related weight in cost function $W_d$ is set to 0.8 and $W_l$ is set to 0.2.

\begin{figure*}[t]
\centering
\includegraphics[scale=0.95]{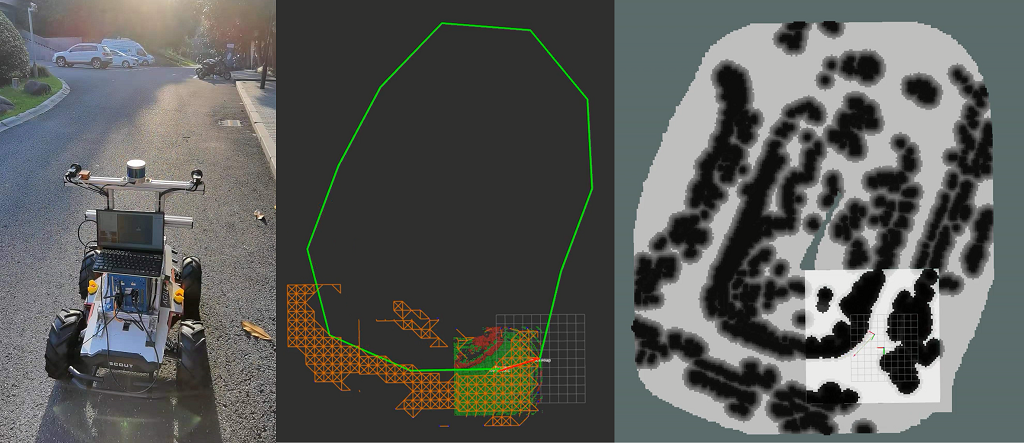}
\caption{A demonstration of the HiTMap in a real-world SLAM system. The left image shows the ground robot we used in a real-world experiment. The middle image is the snapshot when a loop is detected. The right image shows the local and global costmap generated by the modified GEM.}
\label{fig:realWorld}
\vspace{-0.45cm}
\end{figure*}

\begin{figure}
\centering
\includegraphics[scale=0.47]{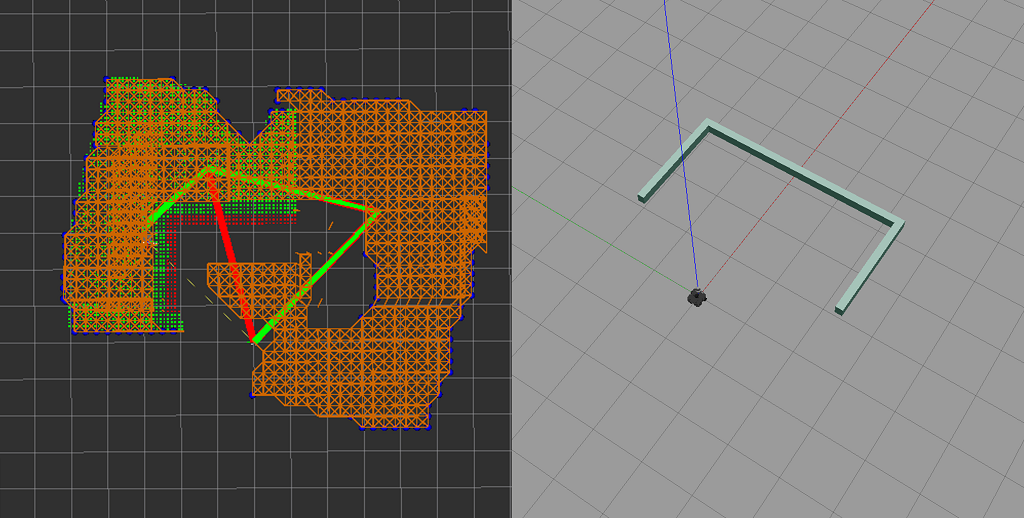}
\caption{A demonstration of the connectivity problem in place recognition.}
\label{fig:prProblem}
\vspace{-0.45cm}
\end{figure}

\subsection{Simulation}

The mobile robot used in the simulation is TurtleBot3 Waffle Pi equipped with an RGB-D camera. We use the TrajectoryPlannerROS as our local planner and the proposed strategy as our global planner.

We first evaluate the HiTMap in a simulation environment as introduced in \cite{tordesillas2019faster}. The room-like scene with a bug trap is hard for navigation but common in indoor applications. Ground robots cannot reach the goal by simply using a local metric map. As shown in Fig. \ref{fig:simulation}, our HiTMap can successfully lead the robot to the goal in such a hard scenario. Two modes of planning are demonstrated in Fig. \ref{fig:twoMode}.

As a map representation, we are more interested in whether it can be used in various scenes, especially large-scale environments. The scalability of the map can be evaluated by the memory usage and calculation time. The performance of the HiTMap and the modified GEM are shown in Fig. \ref{fig:simResult}, the Frame Num means the sensor data sequence, as the sensor collects data in a constant frequency, this number is equal to the time sequence. The curve of GEM is longer in time sequence means the navigation efficiency of GEM is lower than the HiTMap. Also, as we can see from the memory consumption trend in the simulation experiment, although HiTMap occupied more memory for the topological roadmap at the beginning of the evaluation, the long-term memory consumption is bounded. However, the memory usage of the modified GEM grows fastly when the unknown environment is explored. Also, the worst case of modified GEM is shown in the right image of Fig. \ref{fig:simulation}. The back-and-forth movement happens due to the failure of the local planner and the drastic change of global planning. Attribute to the cost function Eq. \ref{eq:utility}, HiTMap successfully avoids the extra movement in the planning process. The time usage trend of comparing methods is shown in Fig. \ref{fig:simResult}. In the small-scale simulation environment without a loop, the advantages of direct correction of the HiTMap are not shown. As we generate roadmap and local metric map in the meanwhile, the time cost by the HiTMap is longer, but it's acceptable.

In the simulation environment, we further design a scenario that can cause the false topology as we mentioned in Section III.C. A low wall is built in the front of the ground robot. The camera can scan a part of the ground behind the wall and thus has the potential of loop closing. When the robot travels around the wall and generates submaps, a loop is detected on different sides of this wall. In the existing topological connection strategy used in aerial robots, an edge is added between loop submaps as shown in Fig. \ref{fig:prProblem}. This edge is useful in pose optimization but harmful in the navigation of ground robots. So with the connectivity check in the HiTMap, this edge is checked and discarded.

\begin{figure}
\vspace{-1.0cm}
\centering
\includegraphics[scale=0.84]{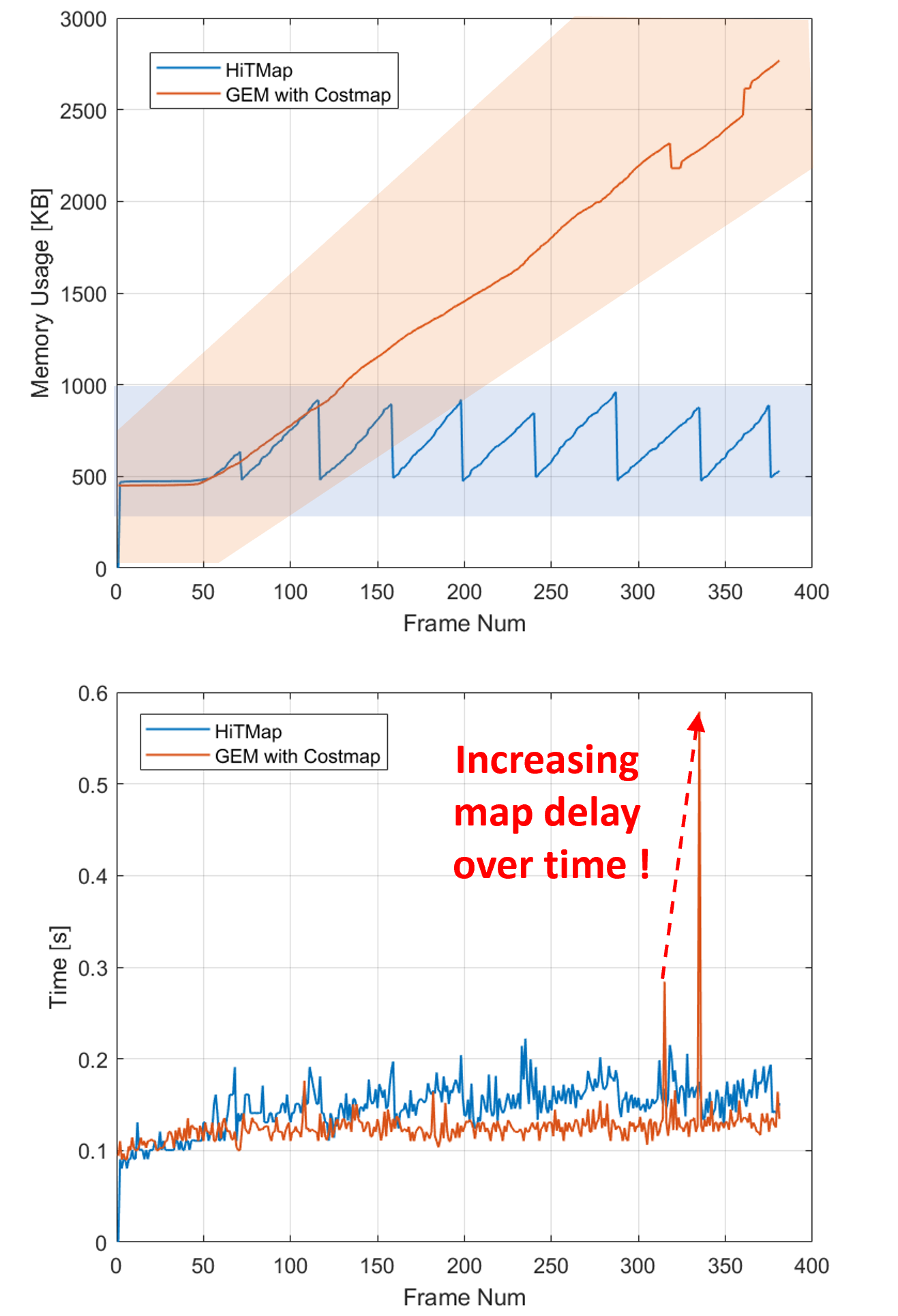}
\caption{The memory usage and time consumption trend in the real-world experiment.}
\label{fig:realResult}
\vspace{-0.35cm}
\end{figure}

\subsection{Real-world Demonstration}

We further test the efficiency of the HiTMap in a real-world SLAM system. The ground robot we used in the real-world experiment is shown on the left of Fig. \ref{fig:realWorld}. It is equipped with several sensors, but we only use lidar to localize, the localization is mainly based on ICP and exhibits drifts during our test. We manually control the robot to move in a loop and validate the correction of topology. The real-world performance on memory usage and time consumption is shown in Fig. \ref{fig:realResult}.

In such a complex and large-scale environment, the different performance on memory usage and time consumption of comparing methods are more significant. As we've analyzed in the simulation experiment, the generation of the roadmap on the local metric map will cost extra time, but the HiTMap doesn't need to costly maintain a global map and thus has bounded time consumption. In this experiment, two loops are detected and triggers the corrections in GEM and leads to an impulse in the time trend curve. The time consumed by this process will increase linearly when exploring and cause delay in path planning.

\section{CONCLUSIONS}

In this paper, we propose the novel hierarchical topological map representation (HiTMap) for navigation in unknown environments. The HiTMap takes advantage of sparse topological representation and the full coverage character of the roadmap. With a navigation strategy implemented, the HiTMap can successfully navigate in unknown environments. With the hierarchical structure, the HiTMap takes bounded time and memory usage in large-scale applications.









\bibliographystyle{IEEEtran}
\bibliography{IEEEabrv,IEEE}

\end{document}